%% file: main.tex
\documentclass[conference]{IEEEtran}
\usepackage{amsmath,amssymb,amsfonts}
\usepackage{algorithmic}
\usepackage{graphicx}
\usepackage{textcomp}
\usepackage{xcolor}

\usepackage{algorithm}
\usepackage{wrapfig}
\usepackage{multirow}
\usepackage{tabularx}
\usepackage{colortbl}
\definecolor{Gray}{gray}{0.9}
\usepackage{array} 
\usepackage{paralist}
\graphicspath{ {figures/} }

\begin{document}

\title{The Potential of Vision-Language Models for Content Moderation of Children's Videos}


\input{authors}
\maketitle
\input{abstract}
\input{introduction}

\input{relatedwork}
\input{method}

\input{results}

\input{conclusion}

\input{bibliography}

\end{document}

%% file: authors.tex
\author{\IEEEauthorblockN{Syed Hammad Ahmed}
\IEEEauthorblockA{\textit{Department of Computer Science} \\
\textit{University of Central Florida}\\
Orlando, USA \\
syed.hammad.ahmed@ucf.edu}
\and
\IEEEauthorblockN{Shengnan Hu}
\IEEEauthorblockA{\textit{Department of Computer Science} \\
\textit{University of Central Florida}\\
Orlando, USA \\
shengnan.hu@ucf.edu}
\and
\IEEEauthorblockN{Gita Sukthankar}
\IEEEauthorblockA{\textit{Department of Computer Science} \\
\textit{University of Central Florida}\\
Orlando, USA \\
gitars@eecs.ucf.edu}
}

%% file: abstract.tex
\begin{abstract}
Natural language supervision has been shown to be effective for zero-shot learning in many computer vision tasks, such as object detection and activity recognition. However, generating informative prompts can be challenging for more subtle tasks, such as video content moderation. This can be difficult, as there are many reasons why a video might be inappropriate, beyond violence and obscenity. For example, scammers may attempt to create junk content that is similar to popular educational videos but with no meaningful information. This paper evaluates the performance of several CLIP variations for content moderation of children's cartoons in both the supervised and zero-shot setting.  We show that our proposed model (Vanilla CLIP with Projection Layer) outperforms previous work conducted on the Malicious or Benign (MOB) benchmark for video content moderation.  This paper presents an in depth analysis of how context-specific language prompts affect content moderation performance. Our results indicate that it is important to include more context in content moderation prompts, particularly for cartoon videos as they are not well represented in the CLIP training data.

\end{abstract}


%% file: introduction.tex
\section{Introduction}
Automated video content moderation is a necessity for protecting children from inappropriate videos uploaded to social media platforms. Every minute YouTube content creators upload around 300 hours of video \cite{ytstats1}. In the United States, more than 80\% children aged 11 years or under watch YouTube videos, out of which almost 50\% have come across inappropriate content as reported by their parents \cite{Auxier_2020}.  Cartoon videos are especially problematic because they are attractive to young viewers.  Scammers often create junk content that includes familiar cartoon characters in order to monetize their videos~\cite{elsa}. Children younger than 6 years of age are more prone to viewing harmful content as they are unable to distinguish between appropriate and inappropriate.  Continuous exposure to malicious video content may have adverse effects on overall cognitive development~\cite{adverse_effect_1}. 

YouTube and other video hosting platforms have tried to mitigate this serious problem of content moderation using automated techniques yet have not been successful in providing robust solutions. YouTube Kids was introduced  as a means to publish child-appropriate videos, but YouTube acknowledges in \cite{ytk_fail} that they are unable to screen all malicious videos. 

Supervised approaches to video content moderation rely on the existence of hand labeled datasets~\cite{sambamulti1,ahmed2023malicious}. 
These datasets are typically small and go out of date quickly as scammers mimic newly trending content.  However, impressive zero shot performance on a variety of computer vision tasks such as object detection~\cite{vanilla_clip}, action recognition~\cite{actionclip}, and depth estimation~\cite{depthclip} has been achieved by joint vision-language models, such as CLIP (Contrastive Language-Image Pre training)~\cite{vanilla_clip}. 
This paper presents an evaluation of the performance of several CLIP variations on the Malicious or Benign benchmark for children's content moderation of cartoon videos~\cite{ahmed2023malicious}.  The Malicious or Benign dataset includes challenging cases of malicious junk content videos that were designed to mimic popular educational cartoons.  These videos lack explicit obscenity or violence but contain loud sounds, fast motions, and characters with scary appearances. Thus it is challenging to devise prompts that are sufficiently general to describe the disturbing activities. Reynolds and McDonell~\cite{reynolds2021promptprogramming} present an overview of prompt programming strategies for large language models that illustrates how different prompting strategies can yield very different outcomes on downstream tasks. 
This paper makes the following contributions:
\begin{compactitem}
    \item Introduces a model (Vanilla CLIP with Projection Layer) that outperforms previous content moderation techniques on the MOB benchmark~\cite{ahmed2023malicious};
    \item Performs an in-depth analysis of how context-specific language prompts affect the content moderation performance of different CLIP variants;
    \item Proposes new benchmark prompt templates for the MOB Dataset.
\end{compactitem}

\begin{figure*}[htp]
    \centering
    \includegraphics[width=0.6\textwidth]
    {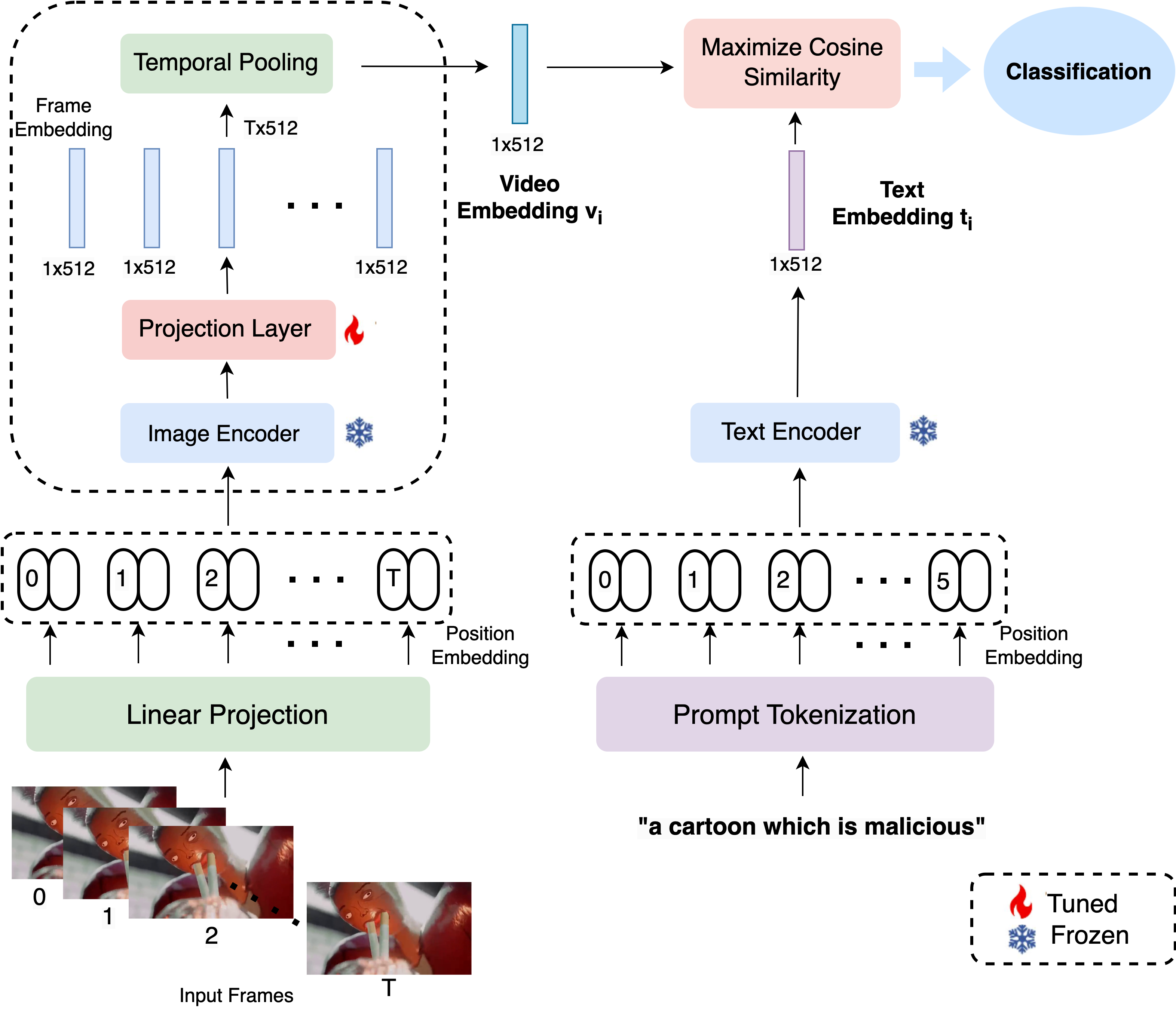}
    \caption{Proposed architecture for language-assisted supervision of video content moderation. The image encoder produces frame-level embeddings with dimensions 1 x 512 after which Temporal Pooling finds the joint representation of all frames of the input video $v_i$,. Similarly, the encoder in the language branch of our model produces text embeddings, $t_i$ after performing tokenization and positional encoding of the input prompt. Finally, cosine similarity is calculated between each ($v_i$, $t_i$) pair which maximizes the similarity for the video for the class for which the related prompt was learnt.}
    \label{fig:model}
\end{figure*}

%% file: relatedwork.tex
\section{Related Work}\label{RW}
Previous work on the application of machine learning for video content moderation for children can be categorized into approaches that focus on a single modality including
meta-data \cite{uni1}, video \cite{taxila, kidsguard}, or text (user comments) \cite{uni3} vs.\ approaches that leverage multimodal data~\cite{sambamulti1,chuttur_multimodal}. The Samba system \cite{sambamulti1} uses subtitles and meta-data to classify inappropriate videos, and Chuttur et al. \cite{chuttur_multimodal} combine both user comments and images to detect inappropriate videos.  

Natural language supervision models such as CLIP~\cite{vanilla_clip} and ALIGN \cite{jia2021scaling_align} combine video and text in a different way than multimodal models.  Rather than using aligned videos and subtitles as the classification input, a natural language prompt is used to describe the instance.  The supervised learning process aims to maximize the similarity between the video embedding and prompt text embedding (Figure~\ref{fig:model}), rather than seeking to minimize classification loss.   Zero-shot video classification approaches extract visual features in the different frames of a video using a backbone such as Convolutional 3D (C3D) \cite{tran2015learning_c3d}, ResNet \cite{he2015deep_resnet} or ViT \cite{VIT}, before learning a temporal behavior that maps visual representations to semantic embeddings.

 Ahmed et al. \cite{ahmed2023malicious} published a hand labeled dataset and a set of video benchmarks for classifying children's cartoon videos as malicious or benign. We demonstrate that our proposed method outperforms these benchmarks on the supervised learning content moderation task.  Unlike other datasets, the MOB dataset includes malicious content that parodies children's educational videos.  The dataset includes a list of audio and video features (summarized in Table \ref{tab:features}) that were used to characterize junk videos.  In this paper we utilize these feature labels for our prompt generation.  

 \begin{table}[h!]
\centering
\caption{Malicious Video and Audio features identified in \cite{ahmed2023malicious} }
\begin{tabular}{|p{0.15\columnwidth} | p{0.75\columnwidth}|}
 \hline
 \textbf{Video} & \textbf{fast repetitive motions, scary/disgusting appearance, hurting/destruction/killing activity, obscene/indecent activity} \\ 
 
 \hline
 \textbf{Audio} & \textbf{loud music/noise, screaming or shouting, explosion or gunshot sounds, offensive language} \\ 
 \hline
\end{tabular}
\label{tab:features}
\end{table}

%% file: method.tex
\section{Methodology}\label{Methodology}
Figure~\ref{fig:model} shows our general architecture for video content moderation.  Videos are classified as malicious or benign based on a short time window of input frames.  We compare the performance of different video embedding and prompt generation strategies for both the supervised and zero shot setting. This paper evaluates the performance of four variants of vision-language pretraining:
\begin{compactenum}
\item \textbf{Vanilla CLIP~\cite{vanilla_clip}} The simplest way to adapt the CLIP model for video classification is to use temporal pooling to create a combined representation of the $T$ frames input to the model. 
\item \textbf{ViFi CLIP~\cite{vifi}} Since CLIP was designed and trained for static images, the temporal representations and object interactions in videos are not explicitly modelled. To address this, ViFi simply performs video fine-tuning on the pre-trained CLIP model. Temporal pooling is used to aggregate representation across frames.
\item \textbf{AIM-CLIP~\cite{yang2023aim}} AIM leverages the CLIP model's video encoders through fully trainable lightweight blocks called \textit{adapters}. Thus, AIM trains on a substantially lower number of parameters. Normally AIM only uses the vision branch of the CLIP model to fine-tune its adapters on video datasets.  In this paper, we incorporated the language transformer in the AIM model and then use it to classify malicious or benign cartoon videos. 
\item \textbf{ActionCLIP~\cite{actionclip}} ActionCLIP was specifically designed for action recognition. It utilizes CLIP for video-text input and makes use of a multi-level prompt strategy.
\end{compactenum}

\subsection{Supervised Learning}
Since full fine-tuning on videos is computationally expensive, we simply add a projection layer on top of the CLIP model while freezing the latter. Projection layers have been used by various models both in fine tuning scenarios and also for dimensionality reduction \cite{projectionlayer1, projectionlayer2, projectionlayer3}.  This helps the model adapt to the downstream task better, as only task-specific parameters are learnt by this layer. Our projection layer has 768 input nodes and 512 output nodes and connects on top of the image encoder which has 768 output nodes. We use ViT \cite{VIT} which uses image patches of size 16 x 16; hence for 3 channels a patch can be fully represented by a vector of size ${16 \times 16 \times 3 = 768}$. We freeze the image and text encoders and leave the projection layer open for tuning as illustrated in Figure \ref{fig:model}.   

\subsection{Zero-shot Classification}
We also performed a zero-shot evaluation on different CLIP-based models. With the exception of Vanilla CLIP, we use respective models pre-trained on the Kinetics-400 dataset \cite{kinetics}. 

\subsection{Prompt Engineering and Ensembling} \label{prompteng}
Modifying the input text prompts can have a significant effect on CLIP's performance~\cite{vanilla_clip,reynolds2021promptprogramming}. After performing rigorous prompt engineering we identified seven prompt templates that empirically performed better than other text prompts during the trial-and-error process. 
\subsubsection{Adding cartoon context to default prompt}\label{context} The standard prompt used in CLIP is \textit{``a photo of a \{ \}."}. As the reference dataset is cartoon only, based on this intuition we added the token \textit{``cartoon"} and devised two new prompts: 1) \textit{``a \{ \} cartoon."} and 2) \textit{``a photo of a \{ \} cartoon."}.
\subsubsection{Prompt candidate generation with cartoon context tokens}\label{promptgen}
A typical prompt used in CLIP is ``a photo of a \{ \}". As discussed in Section \ref{context}, we add the token ``cartoon" so that the tokens become context-specific as the reference dataset is of cartoons only. A general prompt format after adding context becomes: ``a \textit{clip-token} of a \textit{context-token}". From the prompts published for other datasets we enumerated an initial list of clip-tokens and another list of context-tokens from the cartoon synonyms. Based on each prompt generated, we perform zero-shot performance analysis and selected the tokens contributing to the top performing prompts, to create a candidate list of tokens. The initial and final lists are shown in Table \ref{tab:tokens}.


\begin{table}[h!]
\centering
\caption{Initial and Final Candidate Prompt Template token list used to generate context-specific prompts}

\begin{tabular}{|p{0.35\columnwidth} | p{0.55\columnwidth}|}
 \hline
 clip-tokens (initial) & `photo', `video', `example', `demonstration', `image' \\  
 \hline
 context-tokens (initial) & `cartoon', `animation', `caricature', `comic', `character' \\ 
 \hline
 clip-tokens (candidate) & `image', `example'  \\
 \hline
 context-tokens (candidate) & `cartoon', `caricature', `comic' \\
 \hline
\end{tabular}
\label{tab:tokens}
\end{table}


The following formats were used to generate the prompt templates:
\\
\textbf{prompt formats:} 
\\`a \textit{clip-tokens$_{i}$} of a \{\} \textit{context-tokens$_{j}$}.',
\\`a \textit{clip-tokens$_{i}$} of a \textit{context-tokens$_{j}$} which is \{\}.'
\\`a \textit{context-tokens$_{j}$} which is \{\}.'
\\
The prompt templates are generated by placing all pairs (\textit{clip-tokens$_{i}$}, \textit{context-tokens$_{j}$}) in the respective placeholders of each of the three prompt formats.


\subsubsection{Prompt generation using all combinations of cartoon features} \label{featuretokenmethod}
The MOB dataset includes information about the presence or absence of the malicious features shown in Table \ref{tab:features}. Similarly, we devise a list of malicious and benign feature tokens as follows:
\\
\textbf{malicious feature tokens:} ["fast-moving", "scary", "disgusting", "hurting", "destructive", "killing", "obscene", "indecent"]
\\
\textbf{benign feature tokens:} ["good", "friendly", "happy", "joyful", "singing", "enjoying", "loving", "caring", "playing", "funny"]
\\
These lists form a third list of tokens and a candidate list is generated similar to the algorithm described in Section \ref{promptgen}. The prompt format used is:
\\`a \textit{clip-tokens$_{i}$} of a \{\} \textit{context-tokens$_{j}$} which is \{\} and \textit{feature-tokens$_{k}$}.'

\subsubsection{Prompt candidate generation based top-performing frequent token pairs} \label{apriorimethod}
The zero shot evaluation of the exhaustive prompt templates generated from the initial lists described in Section \ref{promptgen} is performed on all pairs (\textit{clip-tokens$_{i}$}, \textit{context-tokens$_{j}$}). The prompt templates with higher than the median performance score (accuracy) are formulated as a list of items in a transaction and given as input to the Apriori rule association algorithm \cite{apriori} which finds the frequent itemsets of size 2. All those frequent pairs become candidate pairs which are then used to generate all prompt templates using the prompt formats discussed in Secion \ref{promptgen}.

\subsubsection{Prompts of other relevant datasets}
We also performed a performance evaluation for the MOB dataset \cite{ahmed2023malicious} using the exact prompt templates as published for other datasets. We evaluated prompts of UCF101 \cite{soomro2012ucf101}, CIFAR10 and CIFAR100 \cite{cifar}, and FER2013 (FacialEmotionRecognition2013) dataset \cite{FER2013}. FER2013 gave the best performance for specific prompt templates for some of the benchmarks.

%% file: results.tex
\section{Evaluation}\label{sec:results}

\begingroup
\renewcommand{\arraystretch}{1.6}
\begin{table*}[!h]
\centering
\caption{\textbf{Supervised setting:} The columns show the different prompts for which we ran our experiments with the benchmarks on the MOB Dataset. The prompts for the frequent item-set combinations outperform other prompts.}

\begin{tabularx}{\textwidth} { 
  >{\raggedright\arraybackslash}X 
  | >{\centering\arraybackslash}X 
  | >{\centering\arraybackslash}X 
  | >{\centering\arraybackslash}X 
  | >{\centering\arraybackslash}X 
  | >{\centering\arraybackslash}X 
  | >{\centering\arraybackslash}X 
  | >{\centering\arraybackslash}X 
  | >{\centering\arraybackslash}X }
 \multirow{2}{5em}{\textbf{Model}} & \multicolumn{7}{c |}{\textbf{Text Prompt Templates}}\\ 
 \cline{2-8} 
 & \textbf{'a photo of a \{\}.'} & \textbf{'a \{\} cartoon.'} & \textbf{'a photo of a \{\} cartoon.'} & \textbf{clip+context token pair combinations} & \textbf{feature-based token combinations} & \textbf{frequent item-set combinations} & \textbf{FER-2013} \\
 \hline
 \hline
    Vanilla (PL) & 78.9 & 78.1 & 77.8 & 76.3 & 75.2 & \textbf{80.3} & 78.5 \\ \hline
    ViFi-CLIP (PL) & 71.0 & 71.3 & 69.9 & 69.2 & 71.3 & \textbf{71.7} & \textbf{71.7} \\ \hline
    AIM-CLIP (PL) & 68.5 & 65.5 & 67.7 & 66.7 & 66.3 & \textbf{69.9} & 67.7 \\ \hline
 
\end{tabularx}

\label{tab:supervised}
\end{table*}
\endgroup

\begingroup
\renewcommand{\arraystretch}{1.6}
\begin{table*}[!h]
\centering
\caption{\textbf{Zero-shot setting: } The prompts using multiple combinations of context tokens give the best zero-shot classification score.}

\begin{tabularx}{\textwidth} { 
  >{\raggedright\arraybackslash}X 
  | >{\centering\arraybackslash}X 
  | >{\centering\arraybackslash}X 
  | >{\centering\arraybackslash}X 
  | >{\centering\arraybackslash}X 
  | >{\centering\arraybackslash}X 
  | >{\centering\arraybackslash}X 
  | >{\centering\arraybackslash}X 
  | >{\centering\arraybackslash}X }
 \multirow{2}{5em}{\textbf{Model}} & \multicolumn{7}{c |}{\textbf{Text Prompt Templates}}\\ 
 \cline{2-8} 
 & \textbf{'a photo of a \{\}.'} & \textbf{'a \{\} cartoon.'} & \textbf{'a photo of a \{\} cartoon.'} & \textbf{clip+context token pair combinations} & \textbf{feature-based token combinations} & \textbf{frequent item-set combinations} & \textbf{FER-2013} \\
 \hline
 \hline
    Vanilla & 62 & 65.6 & 67 & \textbf{68.5} & 67.7 & 67 & 66.3 \\\hline
    ViFi-CLIP & 53.4 & 57.3 & \textbf{58.8} & 58.4 & 58.1 & 53.4 & 56.3 \\\hline
    AIM-CLIP & \textbf{64.2} & 60.9 & \textbf{64.2} & 62.0 & 54.8 & 60.9 & 60.9 \\\hline
    ActionCLIP & 56.6 & 54.8 & 54.8 & 55.6 & 53.8 & \textbf{60.9} & \textbf{60.9} \\ \hline
 
\end{tabularx}
\label{table:zeroshot}
\end{table*}
\endgroup

This section presents results of our evaluation vs.\ the MOB benchmarks~\cite{ahmed2023malicious}. The MOB dataset is annotated with domain-specific features that commonly occur in inappropriate children's cartoon videos. It includes 1875 clips, each 10 seconds in length and frame rate = 25 fps. The dataset has two classes: malicious and benign.

For fair comparison, we used the ViT-B/16 backbone for all CLIP variants discussed in this paper. ViT-B/16 refers to ViT's Base model with an image input resolution of 224 x 224, patch sizes 16 x 16, 12 layers, 12 attention heads, and 12 layers and 8 heads for text input. In the supervised setting, training was performed for 20 epochs with a batch size of 16 and the number of frames set to 16. Adam optimizer with learning rate 1e-4 was used for all experiments. Experiments were run on a 64-bit system with NVIDIA GeForce RTX 3090 GPU and 12th Gen Intel Core i7 CPU with 64 GB of memory.

\subsection{Supervised Learning}
In this section we present results of our proposed supervised language pretraining setting where we use a learnable projection layer to classify videos as malicious or benign. Table \ref{tab:supervised} presents the overview of results for supervised classification with different prompting strategies. We trained and tested three CLIP-based models, out of which the best performing benchmark was our adapted Vanilla CLIP which is labelled with the suffix \textit{(PL)}. ViFi-CLIP (fine-tuned on the Kinetics-400 \cite{kinetics} video dataset) performed the second-best while AIM-CLIP, our adaptation of AIM for CLIP, stood last in terms of accuracy.  The best overall supervised classification results for the MOB Dataset are achieved by our proposed Vanilla-CLIP PL with a testing accuracy of 80.3\% which beat all previously published benchmarks from \cite{ahmed2023malicious}.  Table \ref{tab:newsota} provides a summary of the comparison vs. the other benchmarks. 

\subsection{Prompt Generation Strategies}
The columns in our tables show the results of different prompt generation strategies.
In all our tables, \textit{clip+context token pair combinations} refers to the prompts discussed in Section \ref{promptgen}. \textit{Feature-based token combinations} refers to prompts explained in Section \ref{featuretokenmethod}, and \textit{frequent item-set combinations} refers to text prompt templates generated through the method described in Section \ref{apriorimethod}. Interestingly for all these three benchmarks, the prompt templates generated using the frequent item-set combinations approach give the best results on supervised learning.

We evaluated the zero-shot performance of different prompt generation strategies in Table \ref{table:zeroshot}.  Although there is no single dominant prompting strategy, the  context-based prompts show better performance during zero-shot inference. The best prompt template is the set of prompts generated using both, clip and cartoon context tokens. For both settings, the top results for each benchmark include prompt templates generated by the techniques which use cartoon context-based tokens. 
 This is important since the initial CLIP dataset primarily contains natural images rather than cartoons. 

\begingroup
\begin{table}[h!]
\centering
\caption{Supervised setting vs.\ other benchmarks} 

\begin{tabular}{|c|c|} 
 \hline
 \textbf{Benchmark} & \textbf{Accuracy (\%)} \\ [0.5ex] 
 \hline
    VTN \cite{vtn} & 77.9 \\
 \hline
     I3D \cite{i3d} & 72.1 \\
 \hline
 ConvLSTM \cite{convlstm} & 69.7 \\
 \hline
 \textbf{Vanilla (PL)} & \textbf{80.3} \\
 \hline
 \end{tabular}
\label{tab:newsota}
\end{table}

%% file: conclusion.tex
\section{Conclusion and Future Work}
In this paper we discussed how the problem of video content moderation for children of ages 1-5 years can be addressed using state-of-the-art language supervision techniques. We explored the usage of a foundation model for language pre-training, CLIP (Contrastive Language–Image Pre-training), and performed evaluations on several CLIP variants in supervised and zero-shot settings. In supervised settings, we employed projection layers to improve training on the cartoon video dataset. We also highlighted how language prompts which include context-specific tokens can affect performance on different video classification models. Lastly, we propose the benchmark prompt templates for MOB Dataset for training and evaluating joint vision-language models like CLIP. 

We believe that less well defined video analysis problems such as content moderation pose a significant challenge to prompting strategies. It is challenging for humans to verbally define what makes a video inappropriate. There is a famous quote about pornography by a Supreme Court Justice: ``I shall not today attempt further to define the kinds of material I understand to be embraced within that shorthand description, and perhaps I could never succeed in intelligibly doing so. But I know it when I see it.''~\cite{pornography}  In the future we plan to apply prompt learning approaches where the prompts are learnable parameters, and also explore the performance on cartoon datasets when fine-tuning is done on other pre-trained video datasets.


%% file: bibliography.tex
\bibliographystyle{IEEEtran}
\bibliography{IEEEabrv,references}